\documentclass[twocolumn, switch]{article} 

\usepackage{preprint}

\usepackage{amsmath, amsthm, amssymb, amsfonts}

\usepackage[numbers,square]{natbib}
\usepackage[utf8]{inputenc}	
\usepackage[T1]{fontenc}	
\usepackage{xcolor}		
\usepackage[colorlinks = true,
            linkcolor = purple,
            urlcolor  = blue,
            citecolor = cyan,
            anchorcolor = black]{hyperref}	
\usepackage{booktabs} 		
\usepackage{nicefrac}		
\usepackage{microtype}		
\usepackage{lineno}		
\usepackage{float}			

\usepackage{lipsum}		

\usepackage{newfloat}
\DeclareFloatingEnvironment[name={Supplementary Figure}]{suppfigure}
\usepackage{sidecap}
\sidecaptionvpos{figure}{c}

\usepackage{titlesec}
\titlespacing\section{0pt}{12pt plus 3pt minus 3pt}{1pt plus 1pt minus 1pt}
\titlespacing\subsection{0pt}{10pt plus 3pt minus 3pt}{1pt plus 1pt minus 1pt}
\titlespacing\subsubsection{0pt}{8pt plus 3pt minus 3pt}{1pt plus 1pt minus 1pt}

\title{An Inductive Transfer Learning Approach using Cycle-consistent Adversarial Domain Adaptation with Application to Brain Tumor Segmentation}

\usepackage{xwatermark}

\usepackage{authblk}

\author[1*]{Yuta Tokuoka}
\author[2]{Shuji Suzuki}
\author[2]{Yohei Sugawara}
\affil[1]{School of Fundamental Science and Technology, Graduate School of Science and Technology, Keio University, Japan}
\affil[2]{Preferred Networks, Inc., Japan}
\affil[*]{This work was done while the first author was an intern at Preferred Networks, Inc.}

\begin{document}

\twocolumn[ 
  \begin{@twocolumnfalse} 
  
\maketitle

\begin{abstract}

With recent advances in supervised machine learning for medical image analysis applications, the annotated medical image datasets of various domains are being shared extensively.
Given that the annotation labelling requires medical expertise,  such labels should be applied to as many learning tasks as possible.
However, the multi-modal nature of each annotated image renders it difficult to share the annotation label among diverse tasks.
In this work, we provide an inductive transfer learning (ITL) approach to adopt the annotation label of the source domain datasets to tasks of the target domain datasets using Cycle-GAN based unsupervised domain adaptation (UDA).
To evaluate the applicability of the ITL approach, we adopted the brain tissue annotation label on the source domain dataset of Magnetic Resonance Imaging (MRI) images to the task of brain tumor segmentation on the target domain dataset of MRI.
The results confirm that the segmentation accuracy of brain tumor segmentation improved significantly.
The proposed ITL approach can make significant contribution to the field of medical image analysis, as we develop a fundamental tool to improve and promote various tasks using medical images.


\end{abstract}

\vspace{0.35cm}

  \end{@twocolumnfalse} 
] 



\section{Introduction}

To realize successful medical image analysis using supervised machine learning, the use of large amounts of training data is unavoidable \cite{domingos2012few}.
The number of publicly accessible medical image databases has been steadily increasing\cite{armato2011lung, menze2015multimodal, bakas2017advancing, yan2018deeplesion}.
Accurate annotations of medical images are essential for developing various healthcare applications. Annotation tasks require high level of medical expertise, and hence it is efficient to utilize annotated datasets for other similar tasks.
Wang et al. suggested that there exists a high correlation between lesions occurring in the chest (e.g., Infiltration is often associated with Atelectasis and Effusion) \cite{wang2017chestx}.
From this perspective, medical knowledge annotated to a certain domain dataset is regarded useful for training other tasks using datasets from different domains.
However, given the multi-modal representations for each annotated image, it is challenging to adapt the annotation labels to tasks pertaining other domain datasets.
Supervised transfer learning (STL) is a commonly used method for domain adaptation.
The standard STL approach involves training the model on a source domain (SD) dataset and subsequently fine-tuning the model on a target domain (TD) dataset.
In medical image analysis, several studies have been reported that utilize STL \cite{ghafoorian2017transfer}.
Although fine-tuning has been successfully utilized to improve convergence and training, this approach cannot utilize unlabeled data and does not model the domain shift between SD and TD explicitly.

In recent years, various unsupervised domain adaptation (UDA) methods, such as generative model-based and autoencoder-based models, have been proposed\cite{xia2017w, chen2018semantic, zhang2018task}.
In addition, several studies that utilize Cycle-GAN\cite{zhu2017unpaired}, which is a type of generative adversarial networks (GANs), have been reported in the domain of medical image analysis\cite{zhang2018task, chen2018semantic}.
During the training of Cycle-GAN, the dataset does not need to be paired between each domain.

In most cases, public medical image datasets are basically unpaired.
Unpaired data can be regarded as a dataset wherein images from Computed Tomography (CT) and Magnetic Resonance Imaging (MRI) are respectively imaged to the other patients.
Chen et al. adapted a segmentation model learned using SD chest X-ray images with annotated labels of the lung region; this segmentation model was adapted to the TD chest X-ray images without annotation labels by using Cycle-GAN based UDA\cite{chen2018semantic}.
They showed that it is possible to predict TD annotation labels with high accuracy using a trained SD model.
This method can be used to train the model to distinguish between SD and TD.
Therefore, by using this method, we attempt to explicitly induce SD annotation labels for TD tasks.

\begin{figure}[h]
 \centering
 \includegraphics[width=8cm]{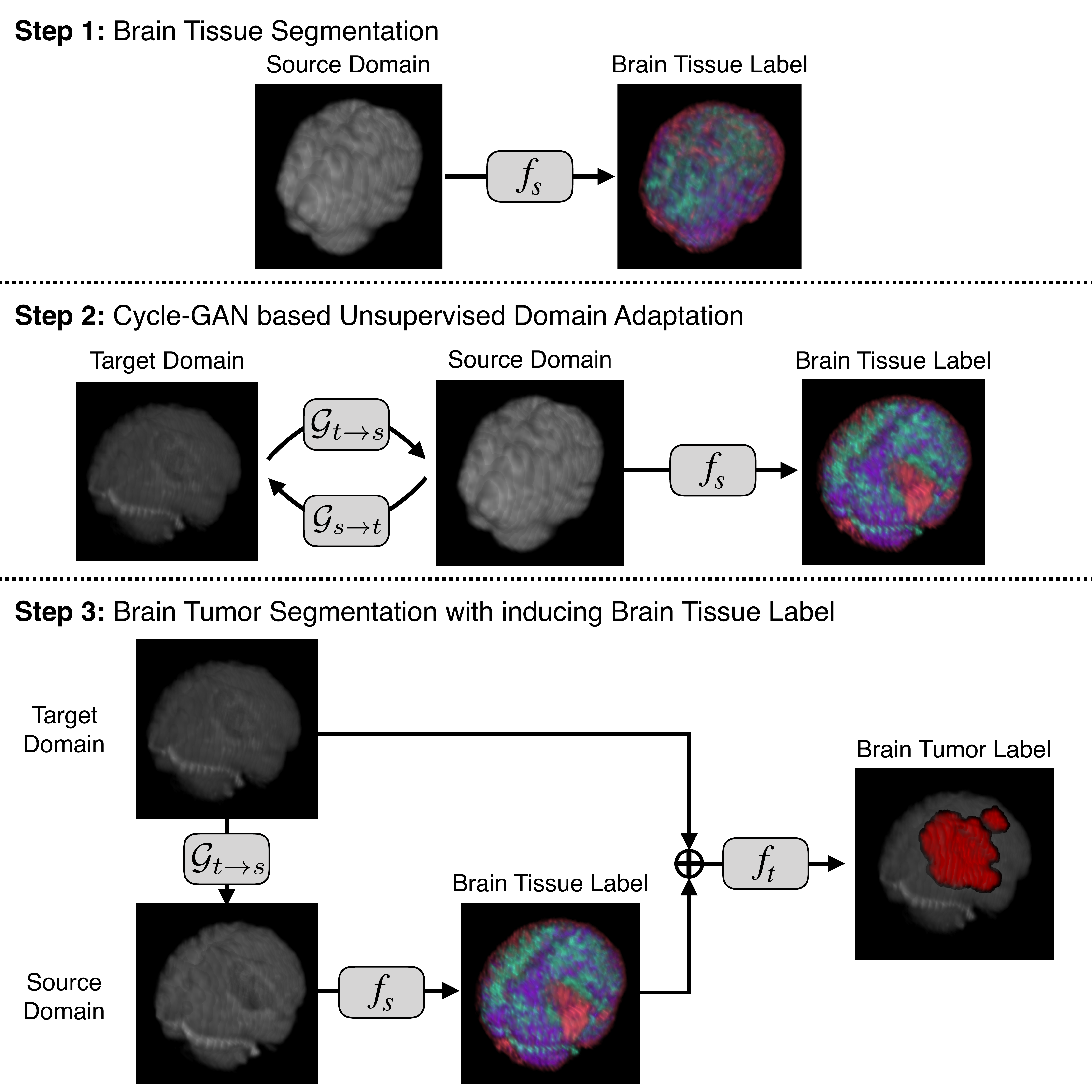}
 \caption{{\bf Overview of the proposed ITL approach for improving brain tumor segmentation.}
 {\bf Step 1:} The segmentor learns to segment the brain tissue using SD dataset;
 {\bf Step 2:} Cycle-GAN based UDA learns to translate TD to SD for adapting the brain tissue segmentation;
 {\bf Step 3:} The TD segmentor learns to segment the brain tumor using induced the annotation label of the brain tissue.
 }
 \label{fig:overview}
\end{figure}

In this study, we propose an inductive transfer learning (ITL) approach wherein the trained SD model extracts annotation labels of the TD dataset by using Cycle-GAN based UDA; these labels are induced for the training task of the TD dataset (Fig. \ref{fig:overview}).
To evaluate the proposed approach, we used MRI dataset, with annotation labels of the brain tumor region, as TD and MRI dataset, with the annotation label of the brain tissue region, as SD.
We demonstrate that the accuracy of the brain tumor segmentation improves through the utilization of brain tissue annotation labels for the training of brain tumor segmentation.

\section{Methods}

The proposed ITL approach induces the annotation label of the brain tissue to learn the brain tumor segmentation (Fig. \ref{fig:overview}).
The approach is sub-divided into three steps; step 1 involves the training phase of the SD segmentor; step 2 is essentially the training phase of UDA; step 3 is the training phase of TD segmentor with inducing annotation label by SD segmentor and UDA.

\subsection{Implementation of Brain Tissue Segmentor}

We implemented the segmentor, which performs the segmentation of the brain tissue based on 3D U-Net\cite{cciccek20163d} (Fig. \ref{fig:architecture}(a)).
We used the softmax function as an output function and Dice loss\cite{milletari2016v} as the objective function.
In general, a commonly occurring problem in the segmentation task, is the presence of imbalanced labels (the number of pixels or voxels in the background and objects).
A previous study reported that it is possible to suppress the influence of dataset label imbalance by using the dice loss function as an objective function\cite{milletari2016v}.

\subsection{Implementation of Cycle-GAN based UDA}

\begin{figure*}[h]
 \centering
 \includegraphics[width=14cm]{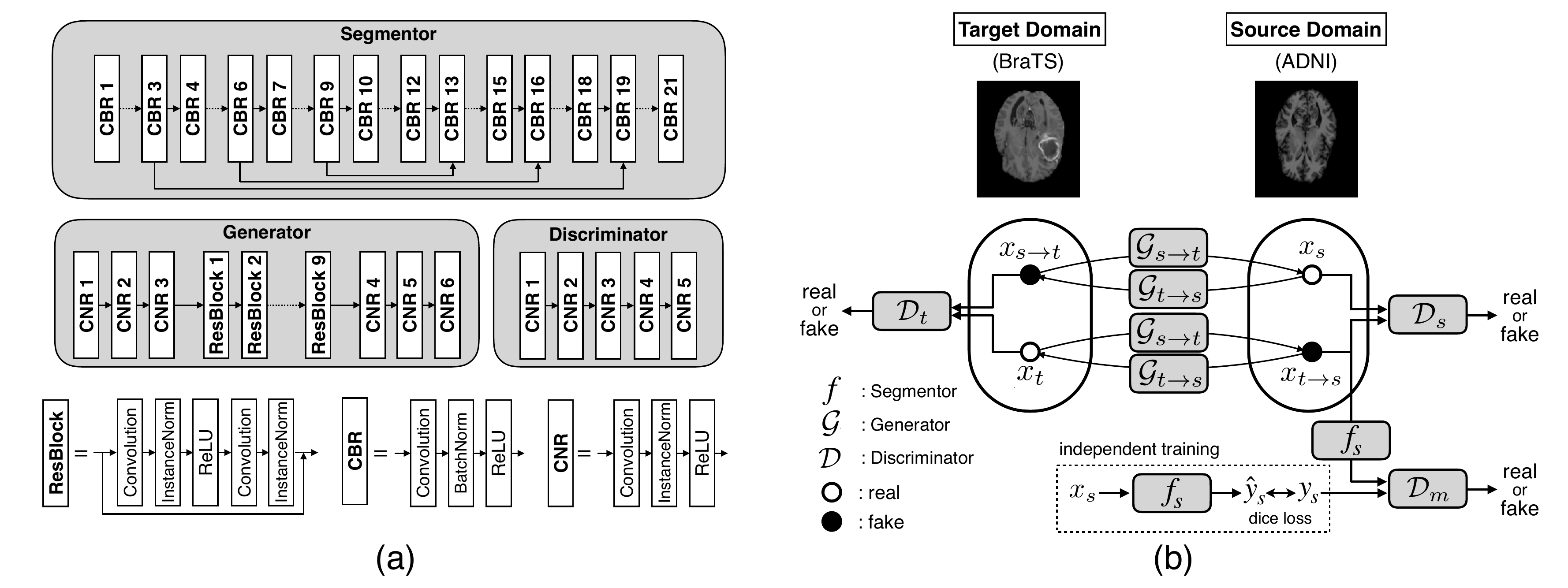}
 \caption{{\bf Overview of architecture in our pipeline.}
 (a) Architecture of segmentor, generator, and discriminator.
 (b) Overview of 3D UDA process.
 }
 \label{fig:architecture}
\end{figure*}

In this work, we implement the Cycle-GAN based UDA for adapting TD to the brain tissue segmentation.
The proposed UDA architecture comprises two generators and three discriminators (Fig. \ref{fig:architecture}(a)) in addition to the trained SD segmentor, which is trained in step 1 (Fig. \ref{fig:architecture}(b)).
We used four types of losses for UDA training.

In general, the purpose of UDA is to obtain the function $f_s$ mapping the SD ($X_s$) to the meaning label space ($Y$) and then adapt the data $x_t$ in the TD ($X_t$) to $f_s$.
In this case, $f_s$ corresponds to the SD segmentor.
To perform unsupervised learning in UDA, the generator $\mathcal{G}_{t \to s}$, which maps data of $X_t$ space to $X_s$ space, and discriminator $\mathcal{D}_s$, which monitors the training result, are required.
The generator generates a fake $x_s$ as $x_{t \to s} = \mathcal{G}_ {t \to s} (x_t)$.
Given that the discriminator distinguishes between the generated false image $x_{t \to s}$ and the real image $x_s$, training of the generator and the discriminator are adversarial in nature.
Therefore, the objective function in this relationship can be defined as follows:
\begin{equation}
 \begin{split}
 \mathcal{L}_{GAN}(\mathcal{G}_{t \to s}, \mathcal{D}_s) &= \mathbb{E}_{x_{s}}[\log \mathcal{D}_s(x_s)] \\
    &\quad+ \mathbb{E}_{x_t}[\log(1 - \mathcal{D}_s(\mathcal{G}_{t \to s}(x_t)))],
 \end{split}
\end{equation}
where the discriminator learns to maximize the objective function and to accurately distinguish between $x_{t \to s}$ and $x_s$.

During image translation, the generated false $x_{t \to s}$ should hold the semantic structure of the real image $x_t$.
We adopted the cycle-consistency loss used in Cycle-GAN\cite{zhu2017unpaired}:
\begin{equation}
 \begin{split}
 \mathcal{L}_{cyc}(\mathcal{G}_{t \to s}, \mathcal{G}_{s \to t}) &= \mathbb{E}_{x_{t}}[||\mathcal{G}_{s \to t}(\mathcal{G}_{t \to s}(x_t)) - x_t||_1] \\
    &\quad+ \mathbb{E}_{x_{s}}[||\mathcal{G}_{t \to s}(\mathcal{G}_{s \to t}(x_s)) - x_s||_1],
 \end{split}
\end{equation}
where the L1$\mathchar`-$Norm is used to reduce blurring in the generated image.
This loss imposes a pixel-wise penalty on the distance between the input image and the cycle converted image.

We also adopted identity loss\cite{taigman2016unsupervised}, which attempts to retain the spatial information within the domain:
\begin{equation}
 \begin{split}
 \mathcal{L}_{ide}(\mathcal{G}_{t \to s}, \mathcal{G}_{s \to t}) &= \mathbb{E}_{x_{s}}[||\mathcal{G}_{s \to t}(x_s) - x_s||_1] \\
    &\quad+ \mathbb{E}_{x_{t}}[||\mathcal{G}_{t \to s}(x_t) - x_t||_1].
 \end{split}
\end{equation}

UDA adapts trained $f_s$ to $x_ {t \to s}$, which is the output of $\mathcal{G}_{t \to s}(x_t)$.
If high quality $x_{t \to s}$ is generated, the accuracy of $f_s(x_{t \to s})$ also improves.
As a constraint for generating higher quality $x_{t \to s}$, we introduce another discriminator $\mathcal{D}_m$ and use semantic-aware loss\cite{chen2018semantic}:
\begin{equation}
 \begin{split}
 \mathcal{L}_{sem}(\mathcal{G}_{t \to s}, \mathcal{D}_m) &= \mathbb{E}_{y_{s}}[\log \mathcal{D}_m(y_s)] \\
 &\quad+ \mathbb{E}_{x_t}[\log(1 - \mathcal{D}_m(f_s(\mathcal{G}_{t \to s}(x_t))))].
 \end{split}
\end{equation}

The full objective function combined with these losses is defined as follows:
\begin{equation}
 \begin{split}
 \mathcal{L}_(\mathcal{G}_{s \to t}, \mathcal{G}_{t \to s}, \mathcal{D}_s, \mathcal{D}_t, \mathcal{D}_m) &= \mathcal{L}_{GAN}(\mathcal{G}_{s \to t}, \mathcal{D}_t) \\
 &\quad+ \alpha \mathcal{L}_{GAN}(\mathcal{G}_{t \to s}, \mathcal{D}_s)  \\
 &\quad+ \beta \mathcal{L}_{cyc}(\mathcal{G}_{t \to s}, \mathcal{G}_{s \to t}) \\
 &\quad+ \gamma \mathcal{L}_{ide}(\mathcal{G}_{t \to s}, \mathcal{G}_{s \to t})  \\
 &\quad+ \epsilon \mathcal{L}_{sem}(\mathcal{G}_{t \to s}, \mathcal{D}_m).
 \end{split}
\end{equation}

In this study, we set parameters $ \{\alpha, \beta, \gamma, \epsilon \}$ as $ \{0.5, 10, 5, 0.5 \}$, respectively.
The overall optimization of UDA was based on generator:
\begin{equation}
 \mathcal{G}^{*}_{s \to t}, \mathcal{G}^{*}_{t \to s} = \arg \min_{\mathcal{G}_{s \to t}, \mathcal{G}_{t \to s}} \max_{\mathcal{D}_s, \mathcal{D}_t, \mathcal{D}_m} \mathcal{L}_(\mathcal{G}_{s \to t}, \mathcal{G}_{t \to s}, \mathcal{D}_s, \mathcal{D}_t, \mathcal{D}_m).
\end{equation}

First, we used Adam as the optimizer with an initial learning rate of 0.002.
Then, we implemented the SD segmentor and UDA based on Chainer, which used NVIDIA Tesla P100 for the calculation of training and inference.

\subsection{Training Procedure of Brain Tissue Segmentor}

In this study, we adopt the top-performing algorithm from the BraTS 2017 challenge as the TD segmentor and baseline model\cite{wang2017automatic}.
We extended the data by concatenating the annotation labels extracted by UDA to the input image.
The annotation label was defined as the probability map of brain tissue segmentation by UDA.
Given that there are four semantic labels of brain tissue segmentation, there are eight channels of input images after concatenation.

\section{Experiments and Results}

\subsection{Datasets}

We used two public brain MRI datasets (BraTS\cite{menze2015multimodal, bakas2017advancing} and ADNI, see \url{www.adni-info.org}) to validate our approach.
The BraTS has four channels for inputs: T1-weighted, T2-weighted, contrast-enhanced T1, and FLAIR.

The BraTS dataset contains images of 484 patients.
We converted all the images to $120 \times 120 \times 77$ voxel by crop and linear interpolation, and scaled the intensity values to $[0, 1]$.
The BraTS dataset is annotated voxel-wise, with four label classes: background, edema, non-enhancing tumor, and enhancing tumor.
We used the three classes for evaluation in the brain tumor segmentation task\cite{menze2015multimodal}.
The whole tumor (WT) is a class including the areas of edema, non-enhancing tumor, and enhancing tumor.
The tumor core (TC) is a class including the area of non-enhancing tumor and enhancing tumor.
The enhance tumor (ET) is the area of enhancing tumor.

As in the case of the BraTS dataset, the ADNI dataset also has four channels.
We annotated the ADNI dataset with brain tissue labels by using an effective approach\cite{gulban2018scalable}; consequently, atlas of white matter (WM), gray matter (GM), and cerebrospinal fluid (CSF) in contrast-enhanced T1 images were generated.
In the pre-processing step, we performed skull-stripping\cite{iglesias2011robust} on the ADNI dataset, given that the BraTS dataset does not include skull images.
The ADNI dataset contains images of 147 patients.
We also converted all the images to $128 \times 128 \times 85$ voxel by using crop and linear interpolation and scaled the intensity values to $[0, 1]$.

\subsection{Evaluation of Brain Tumor Segmentation using ITL}
\begin{figure}[h]
 \centering
 \includegraphics[width=8cm]{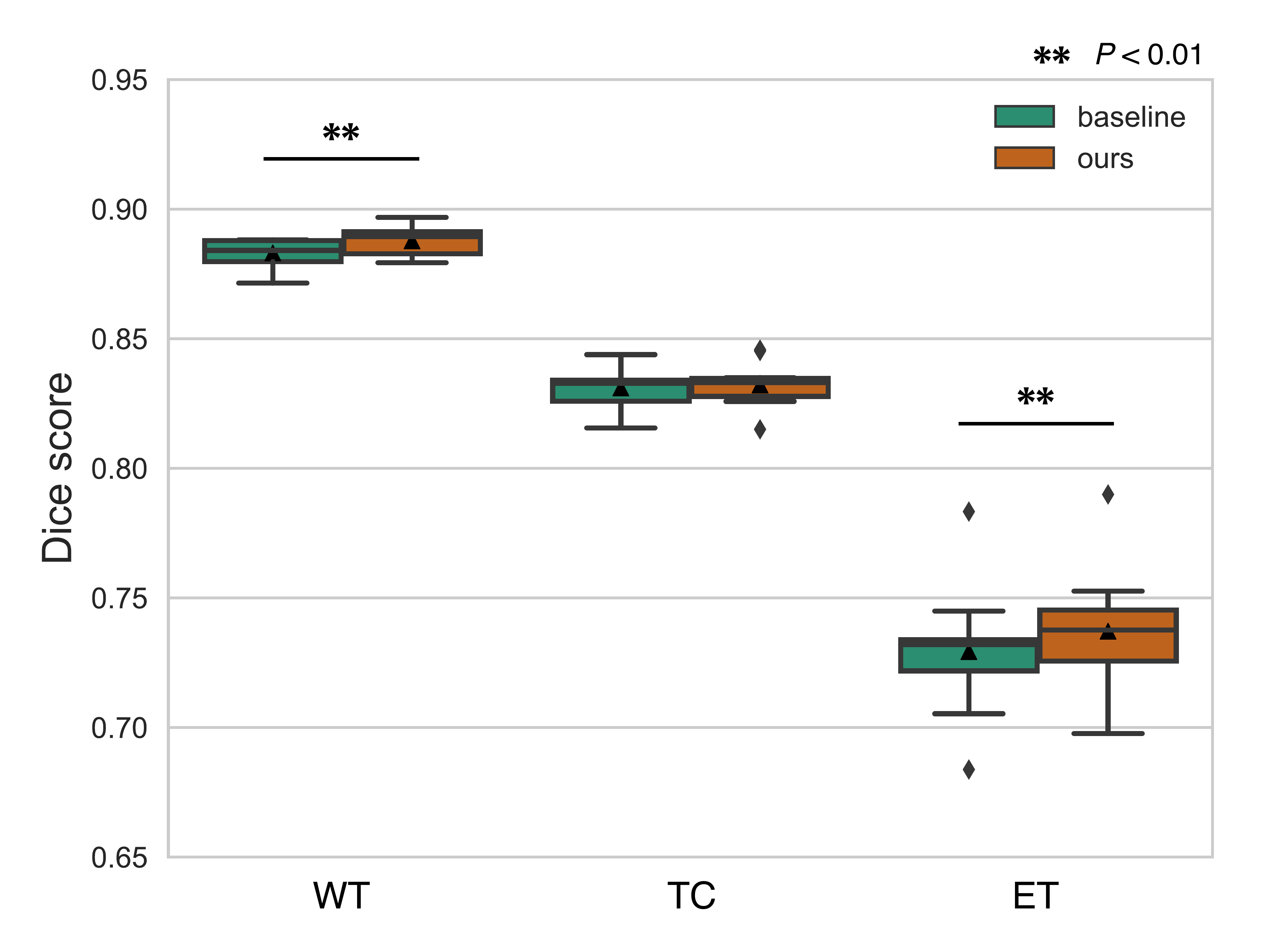}
 \caption{{\bf Quantitative evaluation of annotation transfer.}
 The boxes indicate quartiles including the median; triangles indicate the mean; dots indicate outliers.
 There is a significant difference in the dice score of the proposed method and the baseline ($P < 0.01$, Wilcoxon signed-rank test).
 The mean dice score of baseline WT, TC, and ET is 0.883, 0.831, and 0.729, respectively. On the other hand, the mean dice score for WT, TC, and ET in our proposed method is 0.888, 0.832, and 0.737, respectively.
 }
 \label{fig:annotation_transfer}
\end{figure}

We evaluated the dice score of the brain tumor segmentation task by randomly dividing the dataset into train : validation : test = 7 : 1 : 2 for eleven times in Monte Carlo cross-validation.
As a result of inducing the brain tissue annotation label for tasks of brain tumor segmentation, the mean dice score obtained using the proposed method was superior to that of the baseline (Fig. \ref{fig:annotation_transfer}).
In addition, we performed the paired Wilcoxon rank-sum test and showed that the accuracy was significantly improved, as indicated by the lower p-value score. 
From these results, we demonstrate that brain tissue annotation labels can be useful for brain tumor segmentation.

\begin{figure*}[h]
 \centering
 \includegraphics[width=14cm]{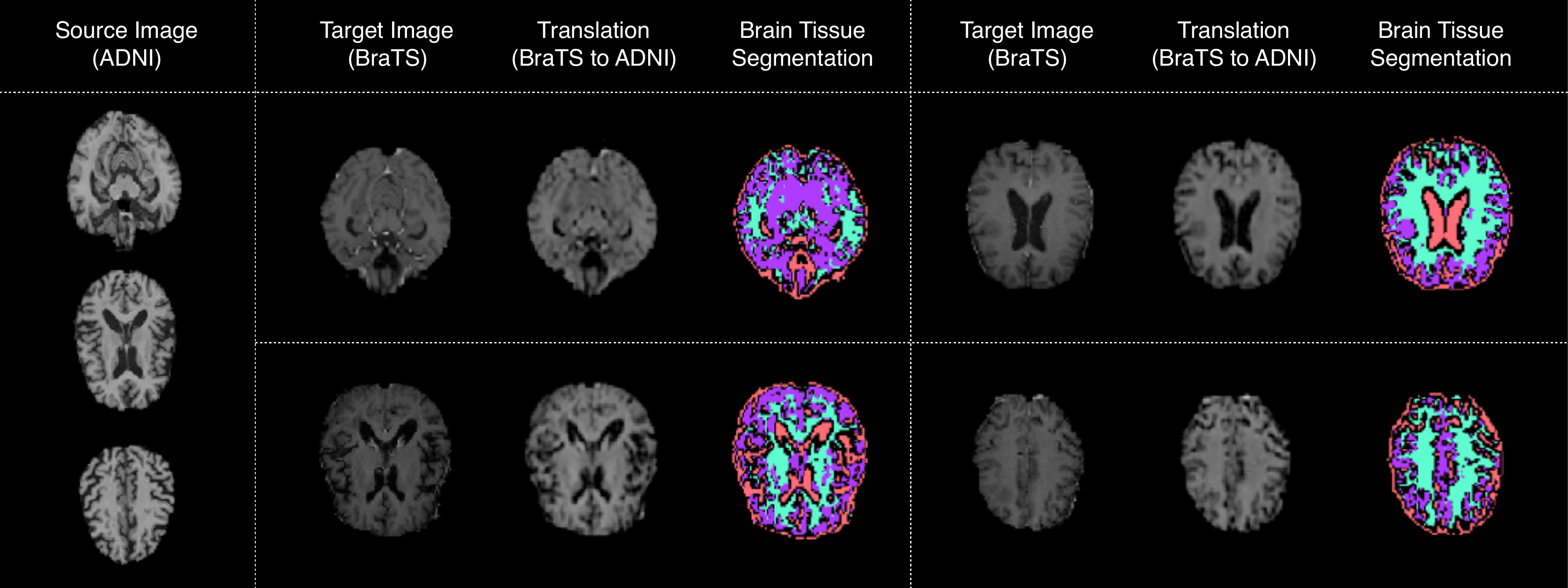}
 \caption{{\bf Visualization of translation and segmentation.}
 Three examples of source images and four examples of target image, translation, and segmentation are shown.
 Green region indicates WM; purple region indicates GM; red region indicates CSF.
 }
 \label{fig:uda_comparison}
\end{figure*}

\subsection{Evaluation of SD Segmentor and Cycle-GAN based UDA}
We evaluated the training results of the SD segmentor and visualized the results of UDA.
We randomly divided ADNI dataset into train : validation : test = 7 : 1 : 2.
For the segmentation task, training was performed using train and validation dataset.
When evaluated using the test dataset, the dice score of each class was 0.7086 (WM), 0.4883 (GM), and 0.6074 (CSF).
Thus, we consider the proposed model to be a suitable SD segmentor for extracting the annotation information of brain tissue.

We trained UDA using the train and validation data in ADNI and BraTS datasets.
We translated the test dataset of BraTS into ADNI domain by using trained generator $\mathcal{G}_x$.
We qualitatively confirmed that the brain tissue, including the brain tumor region, is sufficiently segmented (Fig. \ref{fig:uda_comparison}).
From these results, domain adaptation to the SD segmentor by using UDA can be regarded as successful.

\section{Discussion and Conclusion}
We showed that the annotation labels of the brain tissue can be applied for the training of the segmentation task of the brain tumor (Fig. \ref{fig:annotation_transfer}).
This result also suggests that the formation of the brain tumor is closely associated with the distribution of brain tissue.
We evaluated our ITL approach only using the combination of \{MRI, MRI\}.
However, the evaluation can also utilize combinations of \{CT, MRI\}, \{CT, CT\} and so on.
Moreover, it is possible use combinations of both 2D and 3D images, such as \{X-ray, CT\}, as well as 2D images \{X-ray, X-ray\}.
In this study, we used only one domain that extracts annotations; therefore, our ITL approach can adopt annotation labels of multiple SD to the training of TD.







\normalsize
\bibliography{references}


\end{document}